\documentclass[lettersize,journal]{IEEEtran}

\usepackage{tikz}
\newcommand\copyrighttext{%
  \footnotesize \textcopyright This work has been submitted to the IEEE for possible publication. Copyright may be transferred without notice, after which this version may no longer be accessible.}
\newcommand\copyrightnotice{%
\begin{tikzpicture}[remember picture,overlay]
\node[anchor=south,yshift=10pt] at (current page.south) 
  {\fbox{\parbox{\dimexpr\textwidth-\fboxsep-\fboxrule\relax}{\copyrighttext}}};
\end{tikzpicture}%
}

\usepackage{amsmath,amsfonts}
\usepackage{algorithmic}
\usepackage{algorithm}
\usepackage{array}
\usepackage[caption=false,font=normalsize,labelfont=sf,textfont=sf]{subfig}
\usepackage{textcomp}
\usepackage{stfloats}
\usepackage{url}
\usepackage{verbatim}
\usepackage{graphicx}
\usepackage{cite}
\usepackage{bm}
\usepackage{booktabs}
\usepackage{multirow}
\usepackage{color}
\newcommand{\rev}[1]{#1}
\begin{document}

\title{Deep Active Inference with Diffusion Policy and Multiple Timescale World Model for Real-World Exploration and Navigation}

\author{Riko Yokozawa, Kentaro Fujii, Yuta Nomura, and Shingo Murata,~\IEEEmembership{Member,~IEEE}
\thanks{
This work was supported by the Japan Science and Technology Agency (PRESTO Grant Number JPMJPR22C9) and JSPS KAKENHI Grant Number JP24K03012. (Corresponding author: Shingo Murata)
}%
\thanks{
The authors are with the School of Integrated Design Engineering, Keio University, Yokohama, Kanagawa 223-8522, Japan (e-mail: murata@elec.keio.ac.jp).}}

\markboth{Journal of \LaTeX\ Class Files,~Vol.~18, No.~9, September~2020}%
{How to Use the IEEEtran \LaTeX \ Templates}

\maketitle
\copyrightnotice

\begin{abstract}
Autonomous robotic navigation in real-world environments requires exploration to acquire environmental information as well as goal-directed navigation in order to reach specified targets.
Active inference (AIF) based on the free-energy principle provides a unified framework for these behaviors by minimizing the expected free energy (EFE), thereby combining epistemic and extrinsic values.
To realize this practically, we propose a deep AIF framework that integrates a diffusion policy as the policy model and a multiple timescale recurrent state-space model (MTRSSM) as the world model.
The diffusion policy generates diverse candidate actions while the MTRSSM predicts their long-horizon consequences through latent imagination, enabling action selection that minimizes EFE.
\rev{Real-world navigation experiments, including baseline comparisons, component ablations, and robustness evaluations, demonstrated that our framework achieved higher success rates and fewer collisions, particularly in exploration-demanding scenarios.}
These results highlight how AIF based on EFE minimization can unify exploration and goal-directed navigation in real-world robotic settings.

\end{abstract}

\begin{IEEEkeywords}
Active inference, autonomous navigation, diffusion policy, free-energy principle, mobile robot, world model
\end{IEEEkeywords}

\section{Introduction}

\IEEEPARstart
{A}{utonomous} robotic navigation in real-world environments requires exploration to acquire environmental information as well as goal-directed navigation in order to reach designated targets efficiently.
Achieving an adaptive balance between these two behaviors remains a fundamental challenge in machine learning and robotics \cite{KAELBLING199899, Thrun2005, Levine_2022}.
In many real-world situations, a robot cannot determine its position or the surrounding structure from current observation alone.
For example, in visually similar areas such as corridors or intersections, visually similar observations might correspond to multiple possible locations, creating uncertainty in self-localization \cite{Nowakowski, Aliasing}.
In such cases, exploration plays a crucial role by actively collecting additional information to resolve this uncertainty \cite{Nowakowski, Aliasing, shah2021rapid}.
When sufficient knowledge has been acquired, the robot must shift its focus to goal-directed navigation in order to reach targets efficiently.
Thus, both exploration and navigation are indispensable, and autonomous systems must be able to flexibly balance between them according to the situation.

Traditional approaches such as SLAM-based navigation and handcrafted planners can provide reliable goal-reaching strategies but tend not to generalize to unseen environments or adapt to unexpected situations \cite{SLAM, Kadian2020, Gervet2023}.
Recent advances in learning-based methods, including transformer-based policies \cite{Chen2024, Navformer, ViNT, Lawson2023} and diffusion-based policies \cite{NoMaD, LDP, DD, NavDP}, have enabled diverse action generation and shown promising results in navigation tasks.
However, these methods rely on explicit planners or extensive task-specific supervision to balance exploration and navigation, limiting their flexibility in real-world settings.

Meanwhile, active inference (AIF) based on the free-energy principle (FEP) \cite{FEP} offers a unifying framework for exploration and goal-directed behavior by minimizing the expected free energy (EFE) \cite{AIF, AIFbook, Friston2015}.
EFE comprises two terms: epistemic value, which naturally encourages exploration; and extrinsic value, which accounts for goal-directed behavior.
Prior work has demonstrated the potential of AIF-based navigation in simulation environments \cite{detinguy2024, simAIF}; however, applications in real-world robotic systems remain relatively limited, due mainly to challenges in scaling AIF to complex and uncertain environments.

\begin{figure*}[!t]
    \centering
    \includegraphics[width=1.0\linewidth]{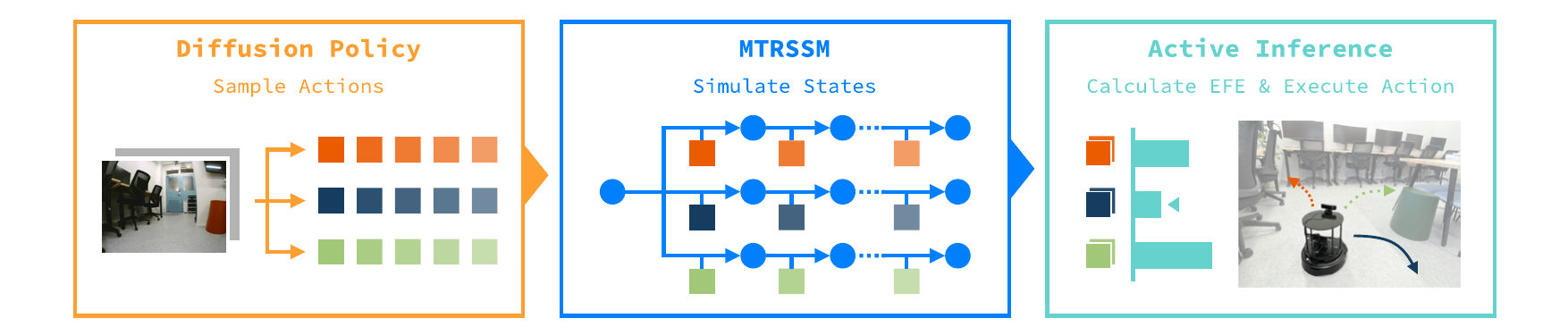}
    \caption{
    Overview of the proposed deep active inference (AIF) framework.
    The framework integrates a diffusion policy and a multiple timescale recurrent state-space model (MTRSSM).
    The diffusion policy generates diverse candidate action sequences, and the MTRSSM predicts the resulting state transitions.
    The expected free energy (EFE) is evaluated for each candidate sequence, and the action with the lowest EFE is selected for execution in the real-world environment.
    }
    \label{fig:overview}
\end{figure*}

Building on this theoretical foundation, we aim to enhance the scalability of AIF for real-world robotic navigation.
Achieving such scalability requires the ability to both generate diverse action sequences depending on the situation and to predict long-horizon state transitions under uncertainty.
These capabilities can be realized by leveraging advances in deep generative models, which offer flexible policy representations and powerful predictive dynamics.
In this work, we propose a deep AIF framework that integrates a diffusion policy \cite{DP} as the policy model and a multiple timescale recurrent state-space model (MTRSSM) \cite{MTRSSM} as the world model \cite{HaWM, masteringatari, Taniguchi03072023} (Fig.~\ref{fig:overview}).
The diffusion policy flexibly generates diverse candidate actions according to the situation \cite{NoMaD}, while the MTRSSM predicts their long-horizon consequences through latent imagination \cite{MTRSSM}. Together, these components enable principled action selection under EFE minimization, thereby balancing exploration and goal-directed navigation without relying on handcrafted planners.

The main contributions of this paper are summarized as follows:
\begin{itemize}
    \item We advance the theoretical foundation of AIF by formalizing how deep generative models can extend the scalability of AIF for real-world navigation, highlighting the role of policy models in generating diverse candidate actions as well as that of world models in supporting long-horizon predictive dynamics.
    \item We realize this approach by using a deep AIF architecture that integrates a diffusion policy as the policy model and an MTRSSM as the world model, thereby enabling principled action selection via EFE minimization while balancing exploration and goal-directed behavior.
    \item \rev{We validate the proposed framework on a real mobile robot, showing that it achieves both exploration and goal-directed navigation in uncertain environments, and further evaluate its robustness through navigation-baseline comparisons, component ablations, and unseen-scenario evaluations.}
    \end{itemize}

The remainder of this paper is structured as follows.
Section II reviews related work on learning-based navigation policies, world models, and AIF.
Section III introduces the proposed methodology, including the formulation of EFE and its integration with a diffusion policy and an MTRSSM.
Section IV describes the experimental setup.
Section V reports the results.
Section VI discusses the findings, and Section VII concludes the paper and outlines future directions.

\section{Related Work}
Research on autonomous navigation has investigated numerous approaches, ranging from policy learning to world models and AIF frameworks.
This section reviews the representative studies in these domains, with a particular focus on their applicability to real-world navigation as well as their limitations.

\subsection{Policy Models}
\rev{Policy learning for navigation has recently advanced along several complementary directions.
Reinforcement-learning-based approaches have been used for fast trajectory planning and control in unknown environments \cite{Chai2024}, while relation-aware policy learning has been investigated for visual navigation \cite{Zhou2024}.}
In parallel, powerful sequence and generative models, such as transformers \cite{transformer, Chen2024, Navformer, ViNT, Lawson2023} and diffusion models \cite{DP, NoMaD, LDP, DD, NavDP}, have increasingly been leveraged for navigation.
Transformer-based methods have demonstrated strong performance in both simulation \cite{Chen2024, Navformer} and real-world environments \cite{ViNT, Lawson2023}, often generating actions by conditioning on subgoals or return signals.
Although these approaches can generalize to real-world scenarios with large-scale datasets, they typically rely on external planners for subgoal specification and are thus less suited for generating exploratory behaviors.

In contrast, diffusion-based policies have been applied to both exploration and navigation, demonstrating robust performance across diverse environments, including real-world settings \cite{NoMaD, LDP, DD, NavDP}.
A representative example is NoMaD, which employs diffusion policy to generate diverse action samples for both exploration and goal-directed navigation within a single policy model.
Although this illustrates the flexibility of diffusion-based policies, NoMaD still requires additional components. Specifically, high-level planners are used to guide action selection during navigation, and the switching between exploration and navigation is determined manually, depending on whether a goal image is provided.
More broadly, balancing these two modes remains a fundamental challenge for diffusion-based methods.

In summary, policy learning for navigation has advanced through transformer- and diffusion-based approaches, but challenges remain in terms of achieving both exploratory and goal-directed behaviors without relying on manually designed modules. 
In the present work, we adopt a diffusion policy as the policy model, motivated by its ability to generate diverse behaviors that can support both exploration and goal-directed navigation within a single policy.
By integrating this policy with AIF, we aim to address the challenge of balancing exploration and navigation in real-world environments.

\subsection{World Models}
World models have been studied as a means to capture environmental dynamics and predict future states without requiring direct interaction with the environment \cite{masteringatari, HaWM, Taniguchi03072023}.
A representative example is the recurrent state-space model (RSSM), which combines deterministic and stochastic latent variables in order to learn compact dynamics.

World models can be leveraged in two main ways.
First, they can facilitate policy learning, where imagined rollouts are used to train policies efficiently \cite{WMs, DayDreamer, sekar2020planning}, \rev{or future scene prediction is used as an auxiliary signal to improve action learning \cite{Surfer}}.
Second, they can support action planning, where imagined rollouts are utilized to evaluate candidate action sequences before execution \cite{LatentPlanning, KAELBLING199899, SparseWM}.

Regarding navigation tasks, world models have been studied for both policy learning \cite{X-mobility, ReCoRe} and action planning \cite{NWM, WMNav}.
For policy learning, some methods incorporate semantic information \cite{X-mobility} or contrastive representation learning \cite{ReCoRe}. 
For action planning, navigation world models employ conditional generative dynamics to imagine trajectories for multiple action candidates and evaluate them by comparing the imagined outcomes with goal observations \cite{NWM}.
Recent work has also developed generative models for autonomous driving, aiming to address large-scale real-world challenges \cite{GAIA-1, drivedreamer}.

Nevertheless, world models face persistent limitations.
Prediction errors accumulate over long horizons, making robust long-horizon imagination difficult, especially in partially observable real-world environments.
\rev{
To mitigate this, models with temporal hierarchies or multiple timescales have been proposed.
For example, the MTRSSM \cite{MTRSSM} captures latent dynamics at both fast and slow timescales, while a memory-augmented world model with multiple timescales improves long-horizon prediction by separately representing long- and short-term memory \cite{Cai2023}.
}

In summary, although world models provide versatility for both learning and planning, their deployment in real-world navigation remains challenged by error accumulation in long-horizon prediction.
In the present work, we leverage a world model within the AIF framework, using it to provide the environmental state predictions required for EFE computation.
To address long-horizon error accumulation, we adopt the MTRSSM, which captures temporal dependencies across multiple timescales, thereby improving long-horizon prediction.

\subsection{Active Inference}
AIF is a biologically inspired framework grounded in the free-energy principle, which explains learning, perception, and action in biological agents \cite{FEP, AIF, AIFbook, Friston2015}.
Under this framework, observations are assumed to be generated by hidden states of the environment, and the agent maintains a generative model to infer these hidden states from observations and select actions accordingly.
While prediction errors on observations (termed ``surprise") are minimized through variational free energy (VFE), actions are selected to minimize EFE, which represents future uncertainty and goal-directed preferences.

By selecting actions that minimize EFE, the agent can effectively integrate both exploration and goal-directed behavior.
In the context of mobile robot navigation, this property enables AIF to serve as a unified framework for handling both exploration and goal-directed navigation, enabling the robot to acquire additional information when necessary and to reach specified targets efficiently.

\rev{This perspective differs from curiosity-driven reinforcement learning, where exploration is typically induced by adding an intrinsic reward bonus to an extrinsic environmental reward \cite{Pathak2017, sekar2020planning}.
Such reward-based approaches often require manually designed reward components, including goal-distance rewards, collision penalties, and exploration bonuses.
In contrast, AIF derives epistemic and extrinsic terms from the EFE formulation, allowing exploration and goal-directed behavior to be balanced within a single objective: minimizing future surprise.}

In machine learning and robotics, deep neural networks have been employed to implement probabilistic representations within AIF, including applications to mobile robot navigation \cite{catal2020, detinguy2024, RobotnavasAIF, Bio, simAIF}.
\rev{Beyond navigation, recent work has also demonstrated the relevance of AIF to adaptive robotic control, such as lifelong gait control \cite{Szadkowski2025}.}
In the context of mobile robot navigation, previous research has tended to focus on either the epistemic (exploration-driven) value or the extrinsic (goal-directed) value in isolation.
For instance, some studies have considered only the epistemic value needed to explore and build topological maps \cite{detinguy2024, RobotnavasAIF}, whereas others have emphasized the extrinsic value needed to achieve navigation objectives in simulation or with real robots \cite{catal2020, simAIF}.
Additionally, hierarchical extensions of state-transition models have been incorporated into AIF\cite{detinguy2024, RobotnavasAIF, Bio, simAIF}.

In summary, despite these advances, most experiments remain confined to simulation, and few studies have simultaneously integrated exploration and navigation in real-world scenarios\cite{RobotnavasAIF, Bio}.
Incorporating powerful policy and world models based on deep generative modeling into AIF might open the door to scaling AIF to more complex real-world environments. 
In the present study, we take a step in this direction by integrating policy and world models into AIF for real-world navigation tasks.

\subsection{Summary}
Recent advances in autonomous navigation span policy learning, world models, and AIF.
Transformer- and diffusion-based policies have improved action generation, with diffusion policies supporting both exploration and navigation.
World models enable dynamics prediction for learning and planning, and hierarchical variants such as MTRSSM have improved long-horizon prediction. 
AIF provides a unified method of exploration and goal-directed behavior by minimizing EFE, but most studies remain limited to performing simulations and emphasize only one mode, limiting real-world applicability.

In contrast, our approach extends AIF to complex real-world navigation by (i) employing a diffusion policy to generate diverse candidate actions that support both exploration and goal-directed navigation, (ii) leveraging an MTRSSM to capture temporal dependencies and provide long-horizon state predictions for EFE computation, and (iii) integrating these components within the AIF framework for principled action selection. 
This combination enables efficient balancing of exploration and goal-directed navigation in uncertain environments.

\section{Methodology}\label{sec:method}
\subsection{Overview}
\begin{figure*}[t]
    \centering
    \includegraphics[width=1.0\linewidth]{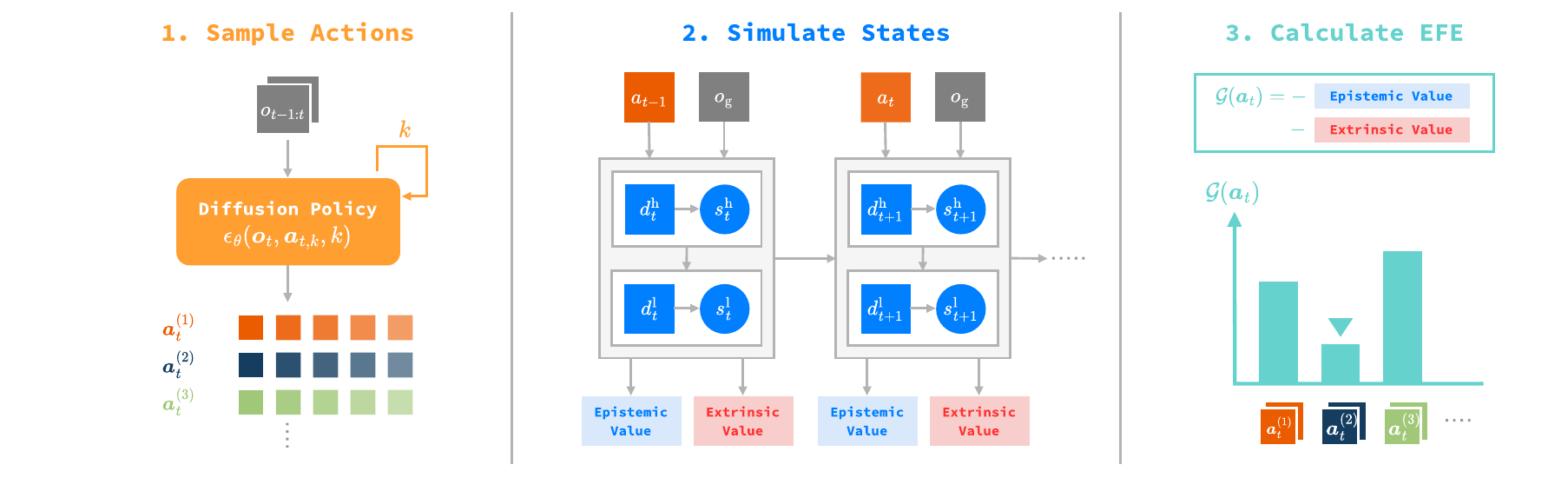}
    \caption{
    Architecture of the proposed deep AIF framework. 
    The process comprises three steps: 
    (1) \textbf{Sample Actions}: the diffusion policy generates multiple candidate action sequences of length $T_{\mathrm{F}}$, conditioned on past observations; 
    (2) \textbf{Simulate States}: the MTRSSM performs latent imagination by simulating the state transitions for each candidate action sequence, sampling high- and low-level latent states in order to estimate epistemic and extrinsic values; 
    (3) \textbf{Calculate EFE}: the EFE is calculated for each candidate action sequence, combining epistemic and extrinsic terms, and the action sequence with the lowest EFE is executed in the real-world environment.}
    \label{fig:architecture}
    \end{figure*}

The proposed deep AIF framework enables autonomous navigation by integrating a diffusion policy as the policy model and the MTRSSM as the world model.
The architecture of the framework is illustrated in Fig.~\ref{fig:architecture}.
The diffusion policy generates diverse candidate action sequences conditioned on past observations, while the MTRSSM predicts the corresponding state transitions.
For each candidate sequence, the EFE is calculated using latent imagination \cite{LatentPlanning}, and the action sequence with the lowest EFE is executed in the real-world environment.

\subsection{Formulation of Active Inference}
In the free-energy principle, perception and learning are formalized as the minimization of VFE. At time $t$, and given hidden states $s_t$ and observations $o_t$ under the generative model $p(o_t,s_t)$, the VFE $\mathcal{F}_t$ serves as an upper bound on the surprise \cite{FEP, stepbystep}:
\begin{align}\label{eq:VFE}
\begin{split}
\mathcal{F}_t&=\mathbb{E}_{q(s_t)}\left[\log{{q(s_t)}-\log{p(o_t,s_t)}}\right]\\
      &=D_{\mathrm{KL}}[q(s_t)||p(s_t)]
      -\mathbb{E}_{q(s_t)}[\log p(o_t|s_t)]\\
      &=D_{\mathrm{KL}}[q(s_t)||p(s_t|o_t)]
      -\log p(o_t)\\
      &\geq-\log{p(o_{t})}.
\end{split}
\end{align}

In contrast, decision-making and action selection are driven by minimization of the EFE.
For a future time step $\tau$ under policy $\pi$, the EFE is defined as follows\cite{stepbystep}:
\begin{equation}
\mathcal{G}_{\tau}(\pi) = \mathbb{E}_{q(o_{\tau},s_{\tau}|\pi)}[\log q(s_{\tau}|\pi) - \log p(o_{\tau},s_{\tau}|\pi)].
\end{equation}
Note that although the VFE is computed after receiving an observation and therefore requires only the expectation over hidden states, the EFE considers future time steps before observations are available, and thus involves expectations over both hidden states and observations.
Following standard derivations, and given preference $C$, the EFE can be decomposed into epistemic and extrinsic values as follows\cite{stepbystep}:
\begin{align}\label{eq:EFE}
\begin{split}
\mathcal{G}_{\tau}(\pi) \approx 
&- \underbrace{\mathbb{E}_{q(o_{\tau}|\pi)}[D_{\mathrm{KL}}[q(s_{\tau}|o_{\tau},\pi)|| q(s_{\tau}|\pi)]]}_{\text{epistemic value}} \\
&- \underbrace{\mathbb{E}_{q(o_{\tau}|\pi)}[\log p(o_{\tau}|C)]}_{\text{extrinsic value}}.
\end{split}
\end{align}

\rev{In AIF, action selection can be formulated as inference over policies. Given a policy $\pi$, the posterior preference over policies is biased toward policies with lower EFE and can be written as: 
\begin{equation}\label{eq:efe_form}
P(\pi) \propto P_0(\pi)\exp[- \mathcal{G}(\pi)],
\end{equation}
where $P_0(\pi)$ denotes a prior over policies. This policy-inference view provides the basis for the action-selection formulation used in this study.}

\subsection{Policy Model}
To enable action selection under AIF, the policy model must represent a diverse set of candidate actions for a given situation.
In this work, we adopt a diffusion policy \cite{DP} as the policy model.
The diffusion policy models the conditional distribution of future action sequences $\bm{a}_t$ based on past observation sequences $\bm{o}_t$, using a diffusion model \cite{dm}; that is, $p(\bm{a}_t|\bm{o}_t)$.
Specifically, we define the action sequence as consisting of the two most recent actions followed by $T_\mathrm{F}$ future steps, and the observation sequence as the two most recent observations:
\begin{equation}\label{eq:seq}
    \left\{ \,
    \begin{aligned}
    &\bm{a}_{t}=a_{t-1:t+T_\mathrm{F}}\\
    &\bm{o}_{t}=o_{t-1:t}
    \end{aligned}
    \right..
\end{equation}

Diffusion models are generative models that are trained to iteratively denoise data corrupted by Gaussian noise.
During training, clean data samples are perturbed with noise, and the model is optimized to predict this noise.
After training, new data can be generated by starting from a noise sample and progressively denoising it through multiple steps.

During training, Gaussian noise is added to the ground-truth action sequence $\bm{a}_t$, and the network $\epsilon_{\theta}$ is optimized to predict the added noise via the following objective:
\begin{align}\label{eq:loss_DP}
\mathcal{L}_{\mathrm{DP}}(\theta) 
&= \mathrm{MSE}(\epsilon_k,\epsilon_{\theta}(\bm{o}_{t},\bm{a}_{t,k},k)), \\
\bm{a}_{t,k} &= \sqrt{\bar{\alpha}_k}\bm{a}_{t,0}+\sqrt{1-\bar{\alpha}_k}\epsilon_k,
\end{align}
where $k$ is the diffusion step, $\epsilon_k \sim \mathcal{N}(0,I)$ is Gaussian noise, and $\bar{\alpha}_k = \prod_{s=1}^k \alpha_s$ denotes the cumulative product of the noise-scheduling coefficients.

At the time of inference, candidate action sequences are generated by iteratively denoising a Gaussian noise sample $\bm{a}^K_t$ through $K$ reverse-diffusion steps, yielding a fully denoised sequence $\bm{a}^0_t$.
The reverse diffusion at step $k$ is defined as:
\begin{equation}\label{eq:DP}
\begin{aligned}
\bm{a}_{t,k-1} = \frac{1}{\sqrt{\alpha_{k}}}\left(\bm{a}_{t,k}-\frac{1-\alpha_k}{\sqrt{1-\bar{\alpha}_k}}\epsilon_{\theta}(\bm{o}_{t},\bm{a}_{t,k},k)\right)+\epsilon_{k},
\end{aligned}
\end{equation}
where $\epsilon_k \sim \mathcal{N}(0,\sigma_{k}^{2}I)$ is Gaussian noise and $\sigma_k$ denotes the standard deviation at step $k$.

At deployment, only the first $T_{\mathrm a}$ steps of the generated sequence $\bm{a}_t$ are executed, while $T_\mathrm{F}>T_{\mathrm a}$ facilitates planning further ahead.

\subsection{World Model}
To predict the action-conditioned state transitions, we used an MTRSSM\cite{MTRSSM}, which extends the standard RSSM\cite{masteringatari} by incorporating temporal hierarchies.
This hierarchical design enables the model to capture both fast and slow dynamics in the environment. 

Let $d^\mathrm{h}_t$ and $d^\mathrm{l}_t$ denote deterministic states at higher and lower levels, respectively, and $s^\mathrm{h}_t$ and $s^\mathrm{l}_t$ denote stochastic states at each level.
The overall latent state becomes
\begin{equation}
    z_t = \{ z^\mathrm{h}_t, z^\mathrm{l}_t \}, \quad z^\mathrm{h}_t = \{ d^\mathrm{h}_t, s^\mathrm{h}_t \}, \quad z^\mathrm{l}_t = \{ d^\mathrm{l}_t, s^\mathrm{l}_t \}.
\end{equation}
The deterministic states evolve according to separate recurrent functions at each timescale as follows:
\begin{align}
    d^\mathrm{h}_t &= f^\mathrm{h}_\phi(d^\mathrm{h}_{t-1}, s^\mathrm{h}_{t-1};\tau^{\mathrm h}), \\
    d^\mathrm{l}_t &= f^\mathrm{l}_\phi(z^\mathrm{l}_{t-1}, s^\mathrm{h}_t, a_{t-1};\tau^{\mathrm l}),
\end{align}
where $f^\mathrm{h}_\phi$ and $f^\mathrm{l}_\phi$ are implemented by multiple timescale recurrent neural networks (MTRNNs) \cite{MTRNN, MT2, murata2017} for the higher and lower levels, respectively. 
The higher level $d^\mathrm{h}_t$ with a larger time constant $\tau^{\mathrm h}$ updates slowly in order to capture long-horizon dependencies, while the lower level $d^\mathrm{l}_t$ with a smaller time constant $\tau^{\mathrm l}$ updates rapidly in order to encode short-term transitions.

The stochastic states are sampled from either the prior distributions $p_\phi^{\mathrm{h}},p_\phi^{\mathrm{l}}$ or the approximate posterior distributions $q_\phi^{\mathrm{h}},q_\phi^{\mathrm{l}}$, defined as follows:
\begin{equation}
    \hat{s}^\mathrm{h}_t \sim p_\phi^{\mathrm{h}}(s^\mathrm{h}_t|d^\mathrm{h}_t), \quad 
    \hat{s}^\mathrm{l}_t \sim p_\phi^{\mathrm{l}}(s^\mathrm{l}_t|d^\mathrm{l}_t),
\end{equation}
\begin{equation}
    s^\mathrm{h}_t \sim q_\phi^{\mathrm{h}}(s^\mathrm{h}_t|d^\mathrm{h}_t, d^\mathrm{l}_t), \quad
    s^\mathrm{l}_t \sim q_\phi^{\mathrm{l}}(s^\mathrm{l}_t|d^\mathrm{l}_t, o_t).
\end{equation}

Observations $o_t$ are encoded into low-dimensional features via a convolutional neural network (CNN) encoder before being fed into the model.
The MTRSSM is trained by minimizing the following VFE-based loss:
\begin{equation}
\begin{aligned}
\mathcal{L}_{\mathrm{WM}} 
= \sum_{t=1}^{T} \Big\{ 
&  \beta D_{\mathrm{KL}}\!\left[q_\phi^{\mathrm{l}}(s^\mathrm{l}_t \mid d^\mathrm{l}_t,o_t) \,\big\|\, p_\phi^{\mathrm{l}}(s^\mathrm{l}_t \mid d^\mathrm{l}_t)\right] \\
+ & \beta D_{\mathrm{KL}}\!\left[q_\phi^{\mathrm{h}}(s^\mathrm{h}_t \mid d^\mathrm{h}_t,d^\mathrm{l}_t) \,\big\|\, p_\phi^{\mathrm{h}}(s^\mathrm{h}_t \mid d^\mathrm{h}_t)\right] \\
- & \mathbb{E}_{q_\phi(s^\mathrm{l}_t|d^\mathrm{l}_t, o_t)}\!\left[\log p_\phi^{\mathrm{l}}(o_t \mid z^\mathrm{h}_t, z^\mathrm{l}_t)\right] \\
- & \mathbb{E}_{q_\phi(s^\mathrm{h}_t|d^\mathrm{h}_t, d^\mathrm{l}_t)}\!\left[\log p_\phi^{\mathrm{h}}(d^\mathrm{l}_t \mid z^\mathrm{h}_{t-1})\right] 
\Big\}.
\end{aligned}
\end{equation}
This loss combines the KL regularization at both hierarchies with the reconstruction losses of observations and dynamics, making it possible for the model to learn both short- and long-horizon dependencies.

\subsection{Active Inference with EFE Minimization}
\begin{figure}
    \centering
    \includegraphics[width=1.0\linewidth]{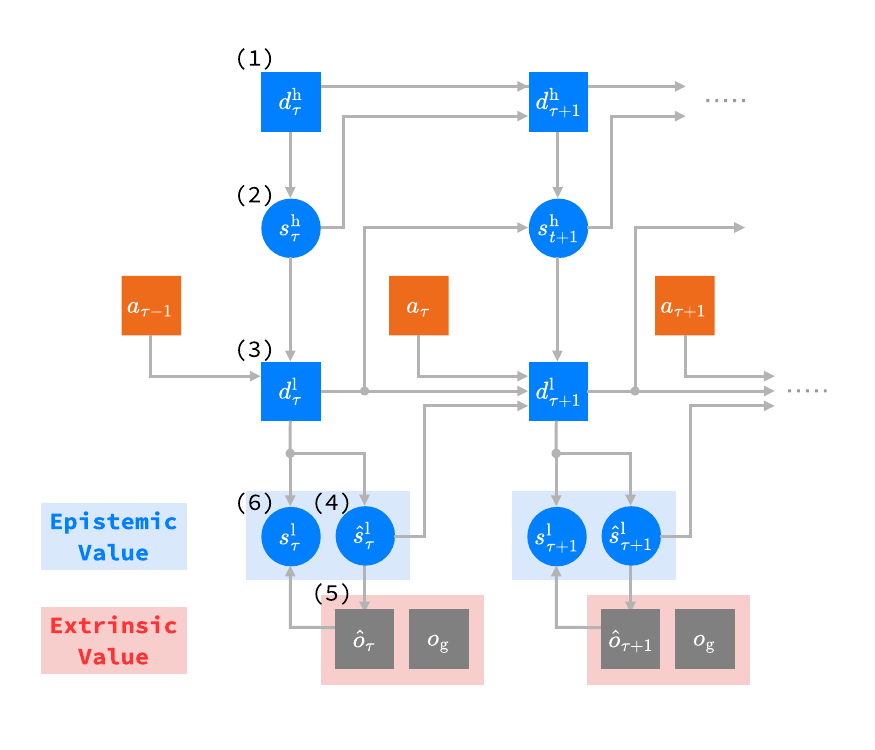}
    \caption{
    Computation of EFE for candidate action sequences.
    Each candidate action sequence $\bm{a}_t$ is simulated by the MTRSSM, using latent imagination.
    At each time step, the process unfolds as follows:
    (1) the higher-level deterministic state $d^\mathrm{h}_{\tau}$ is updated;
    (2) the higher-level stochastic state $s^\mathrm{h}_{\tau}$ is sampled;
    (3) the lower-level deterministic state $d^\mathrm{l}_{\tau}$ is updated;
    (4) the lower-level prior \rev{$p_\phi(s^\mathrm{l}_\tau \mid d^\mathrm{l}_\tau)$} is predicted and the stochastic state $\hat{s}^\mathrm{l}_{\tau}$ is sampled;
    (5) a predicted observation $\hat{o}_\tau$ is generated; and
    (6) the lower-level stochastic posterior $q_\phi(s^\mathrm{l}_\tau \mid d^\mathrm{l}_\tau, \hat{o}_\tau)$ is inferred using $\hat{o}_\tau$.
    At the lower level, the epistemic value is computed as the KL divergence between the posterior and the prior, while the extrinsic value is computed as the feature-space distance between the predicted observation \rev{$\hat{o}_\tau$} and the goal observation $o_\mathrm{g}$.
    Combining these two terms yields the EFE $\mathcal{G}_\tau(\bm{a}_t)$, which is used to select the action sequence that balances exploration and goal-directed navigation.
    }
    \label{fig:EFE}
\end{figure}

At the time of inference, the policy model generates multiple candidate action sequences $\bm{a}_t$, each of which is simulated using the MTRSSM through latent imagination \cite{LatentPlanning}. 
Future states $\hat{s}^\mathrm{l}_\tau$ are sampled from the prior distribution, and the corresponding predicted observations $\hat{o}_\tau$ are decoded. 
At each time step, $M$ high-level latent states $s^\mathrm{h}_\tau$ are sampled from the posterior distribution of the higher level, and for each one, $N$ low-level latent states \rev{$\hat{s}^\mathrm{l}_\tau$} are sampled from the prior of the lower level. 
This results in $M \times N$ predicted observations $\hat{o}^{i,j}_\tau$. 
Using these predicted observations together with the candidate action and the current state, the posterior of the lower level is recomputed at each time step and propagated forward through the MTRSSM. 
The KL divergence between this posterior and the corresponding prior constitutes the epistemic value term.
The EFE for a candidate action sequence is approximated as follows:
\begin{equation}\label{eq:EFE_sample}
\begin{aligned}
\mathcal{G}_\tau(\bm{a}_t) 
&\approx - \frac{1}{MN} \sum_{i=1}^M \sum_{j=1}^N 
\Biggl\{ \\
    &D_{\mathrm{KL}}\!\Bigl[
        q_\phi^\mathrm{l}(s^\mathrm{l}_\tau \mid d^\mathrm{l}_\tau,\hat{o}^{i,j}_\tau) 
        \,\big\|\, 
        p_\phi^{\mathrm{l}}(s^\mathrm{l}_\tau \mid d^\mathrm{l}_\tau)
    \Bigr] \\
&\quad \rev{-} \frac{1}{\sigma_\tau^2} \,
    \mathrm{MSE}\!\bigl(f(\hat{o}^{i,j}_\tau), f(o_{\mathrm{g}})\bigr)
\Biggr\},
\end{aligned}
\end{equation}
where $f(\cdot)$ denotes a CNN encoder that maps observations onto a feature space.
The first term corresponds to the epistemic value, and the second term corresponds to the extrinsic value in (\ref{eq:EFE}). 
Details of the EFE computation are illustrated in Fig.~\ref{fig:EFE}.

In this work, the epistemic value is computed only at the lower level of the MTRSSM, while the extrinsic value is defined as the temporal average of a feature-space distance between the predicted (imagined) observations $\hat{o}_\tau$ and the goal observations $o_{\mathrm g}$. 
To balance these two terms, the precision (inverse variance) $1/\sigma_{\tau}^{2}$ is designed as a time-decaying coefficient, where the epistemic value dominates in the earlier phase when self-localization uncertainty is high, while the extrinsic value gradually becomes dominant in the later phase once the robot has localized itself.

Finally, an action sequence $\bm{a}_t^*$ is selected according to the following equation:
\begin{equation}
\bm{a}_{t}^*= \arg\min_{\bm{a}_t}\mathcal{G}_\tau(\bm{a}_t).
\end{equation}
This action-selection mechanism based on the EFE minimization enables the robot to balance exploration driven by epistemic value with goal-directed navigation guided by extrinsic value.

\rev{
Following the policy-inference formulation introduced in (\ref{eq:efe_form}), we can interpret this procedure as a sampling-based approximation of policy selection under AIF. Specifically, a future action sequence $\bm{a}_t$ is treated as a finite-horizon policy, and the diffusion policy parameterizes a data-driven conditional prior $P_0(\bm{a}_t \mid \bm{o}_t)$ over feasible action sequences.
The MTRSSM provides the predictive distribution required to evaluate $\mathcal{G}(\bm{a}_t)$, yielding:
\begin{equation}
    P(\bm{a}_t \mid \bm{o}_t) \propto P_0(\bm{a}_t \mid \bm{o}_t)\exp[-\mathcal{G}(\bm{a}_t)].
\end{equation}
In practice, rather than performing exact posterior inference, we sample a finite set of candidate action sequences from the diffusion policy and select the sequence with the lowest EFE.
}

\section{Experiments}\label{sec:experiment}
\subsection{Hardware Setup}
The proposed deep AIF framework was implemented with a TurtleBot 4 (Clearpath Robotics), as shown in Fig.~\ref{fig:overview}. The robot is controlled via two velocity commands: linear velocity and angular velocity. A wide-angle RGB camera (CMS-V43BK, Sanwa Supply) was mounted on the top plate of the robot. The captured images were resized to $240\times 320$ pixels and used as observations.

\subsection{Experimental Environment}
\begin{figure*}
    \centering
    \includegraphics[width=1.0\linewidth]{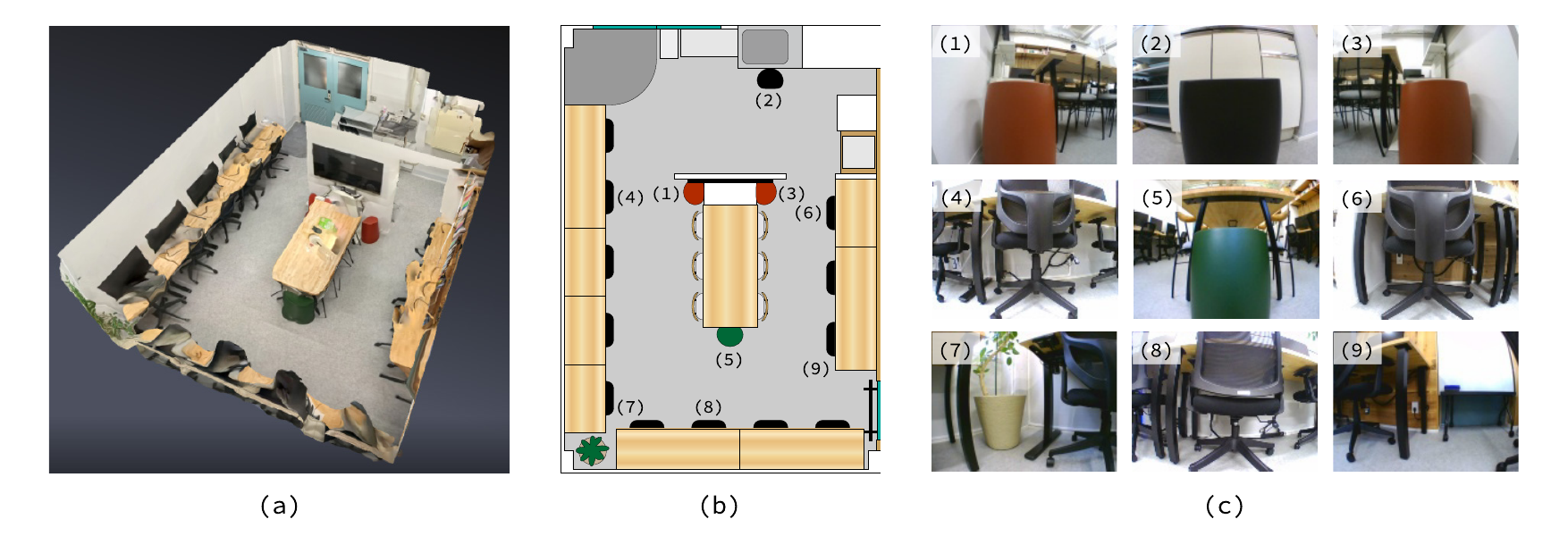}
    \caption{
    Experimental environment.
    (a) Overhead view of the indoor room.
    (b) Top--down map of the environment.
    (c) Representative observations at nine designated location--orientation patterns (1)--(9) on the map.
    In the real-world navigation tasks, the initial position--orientation pairs and goal images were selected from among these patterns, resulting in 18 experimental cases.
    }
    \label{fig:env}
\end{figure*}

The experiments were conducted in an indoor room (approximately 4.7 m $\times$ 6.3 m), the layout and representative observation images of which are shown in Fig.~\ref{fig:env}.
Three sides of the room (excluding the entrance wall) were lined with desks and black desk chairs, and the center contained a meeting table surrounded by dining chairs.
Additionally, four colored stools were placed as landmarks: two red, one black, and one green.
A key challenge is that, because of the similarity of images across different locations, especially along the wall side with black desk chairs, localization based solely on the current observation can be ambiguous.
Therefore, the robot must actively explore the environment to reduce uncertainty and realize goal-directed navigation.

\subsection{Data Collection}
Data were collected by manually teleoperating the robot within the environment. In total, 15 sequences of 2,000 time steps each were recorded at 5 Hz, for a total of 30,000 time steps. Each dataset contained velocity commands and RGB observation images.
\rev{The training dataset consisted of unguided roaming trajectories collected to cover the experimental environment as uniformly as possible, without specifying goal images.}

\subsection{Dataset for Policy Model Training}
For policy model training, the observation images were resized to $120\times 160$ pixels. Each sequence was segmented into 128-step windows every 32 steps, yielding 885 subsequences. From each, 64-step subsequences were randomly sampled to ensure that all observation images would appear as conditions during training.

\subsection{Dataset for World Model Training}
For world model training, the observation images were downsampled to $60\times 80$ pixels. Each sequence was segmented into 600-step trajectories every 100 steps, resulting in 225 trajectories. From each trajectory, 500-step subsequences were randomly extracted.

\subsection{Implementation Details}
\subsubsection{Training Details}
\paragraph{Diffusion policy}
The observation encoder was composed of a 3-layer CNN followed by a spatial softmax, producing 32 keypoints that conditioned the diffusion model.
The CNN used convolutional layers with channels (8, 16, 32), kernel sizes (3, 3, 3), strides (2, 2, 1), and paddings (1, 1, 1).
The diffusion model was implemented as a 1D-UNet \cite{UNet} with a DDIM sampler \cite{DDIM}.
The action sequence length was set to $T_\mathrm{F}=64$, and the number of diffusion steps was set to $K=100$.

\paragraph{MTRSSM}
The observation encoder consisted of a 3-layer CNN with channel sizes (16, 32, 64), kernel sizes (3, 3, 3), strides (1, 2, 2), and paddings (1, 1, 1), producing a 128-dimensional embedding.
The deterministic states of the MTRSSM were modeled using an MTRNN \cite{MTRNN, MT2, murata2017} with both higher-level deterministic states $d_t^\mathrm{h} \in \mathbb{R}^{32}$ and lower-level deterministic states $d_t^\mathrm{l} \in \mathbb{R}^{128}$.
The time constants were set to $\tau^\mathrm{h}=64$ for the higher level and $\tau^\mathrm{l}=4$ for the lower level.
The higher-level stochastic states $s_t^\mathrm{h}$ were represented by a categorical distribution over $4 \times 4$ classes, while the lower-level stochastic states $s_t^\mathrm{l}$ were represented by a categorical distribution over $8 \times 8$ classes.
The image decoder was implemented as a CNN with eight residual blocks and two pixel-shuffle layers \cite{resnet, ps}.
Training employed truncated backpropagation through time \cite{TBPTT} with a window size of 50 steps.

\subsubsection{Experiment Details}
\paragraph{Diffusion Policy}
At test time, the policy generated action sequences of $T_\mathrm{F}=64$ steps but executed only the first $T_\mathrm{a}=32$ steps.
The number of diffusion steps was reduced to $K=10$ for real-time execution, and eight candidate action sequences were sampled per inference.

\paragraph{MTRSSM}
For EFE computation defined in (\ref{eq:EFE_sample}), we sampled $M=5$ higher-level latent states and $N=5$ lower-level latent states at each time step.
The precision for the extrinsic value in (\ref{eq:EFE_sample}) was formulated as a smooth, sigmoidal function of the inference iteration $n$, which was incremented once every $T_{\mathrm a}$ time steps.
Specifically, it was defined as follows:
\begin{equation}\label{eq:precision}
\frac{1}{\sigma_{\tau}^{2}} = 0.08 + \frac{3.0 - 0.08}{1 + \exp[-0.6 (n - 10)]}.
\end{equation}
This formulation gradually shifts the weighting between epistemic and extrinsic values, with the epistemic term dominating in the early phase, and the extrinsic term becoming increasingly influential as the inference iteration progresses.
The feature encoder $f(\cdot)$ comprised a three-layer CNN with channel sizes (16, 32, 64), kernel sizes (3, 3, 3), strides (2, 2, 2), and paddings (1, 1, 1), followed by a three-layer fully connected network with 1,024 hidden units.
The feature encoder was trained so that the feature-space distance between two observations reflects their spatial distance, as estimated from accumulated actions or odometry.

\subsection{Navigation Task with Real-World Robot}\label{sec:navtest}
The main evaluation was conducted using robot navigation experiments in the environment described above.
Three initial positions were prepared, and each was tested under the following two facing directions:  
(1) facing toward the interior of the room, where the current location is visually distinct; and
(2) facing the wall side, where the similarity of chairs along the three walls introduces perceptual aliasing, making localization highly uncertain.
Accordingly, six distinct initial states were defined in total.  

For each initial position, three different goal images were specified, yielding $6 \times 3 = 18$ task cases. Each case was executed twice, for a total of 36 trials. A trial was terminated after 1,000 time steps if the robot had not reached the goal. Success was defined as entering a predefined goal area (manually determined to correspond to the region where the goal image was observable). In addition to the success rate, the number of collisions with obstacles was also recorded.

\rev{
The robot was controlled at 10 Hz, and inference was performed on an NVIDIA RTX 3090 GPU. In the current implementation, action selection required approximately 5 s per decision step, primarily due to EFE computation over candidate action sequences.
}

\subsection{\rev{Baseline Comparison}}
\begingroup\color{black}
The first set of experiments compares the proposed framework with a goal-directed internal baseline and external navigation baselines.

\paragraph{Only Extrinsic}
Only Extrinsic minimizes only the extrinsic component of the EFE, ignoring the epistemic value.
This corresponds to purely goal-directed navigation without exploration, similar to conventional planning- or reward-driven methods.

\paragraph{NoMaD and FlowNav}
We also evaluated NoMaD \cite{NoMaD} and FlowNav \cite{FlowNav} using their publicly released pretrained models.
Both methods output eight waypoints at each inference step.
In our experiments, the predicted waypoints were rescaled using trajectory statistics from the experimental environment, and the first two waypoints were tracked over 32 control steps.
For NoMaD and FlowNav, we collected expert start-to-goal trajectories in the same environment for all 18 task cases and used them to construct the corresponding topological maps.
Note that the topological maps provide NoMaD and FlowNav with additional environment-specific guidance from expert trajectories, whereas the proposed method does not use such task-specific topological maps during evaluation.

\endgroup

\subsection{\rev{Policy and World Model Ablations}}
\begingroup\color{black}
The second set of experiments evaluates the contribution of the major components in the proposed framework.

\paragraph{RSSM}
\rev{The RSSM ablation replaces the MTRSSM with a standard RSSM.}

\paragraph{Gaussian Sampling Policy}
The Gaussian sampling policy (GSP) replaces the diffusion policy with a stochastic Gaussian policy.
The same observation encoder as the diffusion policy was used, and an MLP predicted the mean and variance of future velocity actions.
Candidate action sequences were then sampled from the resulting Gaussian distribution.

\paragraph{Cross-Entropy Method}
The cross-entropy method (CEM) baseline directly optimizes velocity action sequences through iterative sampling and refitting.
At each iteration, 50 candidate sequences were sampled, and the top 10\% with the lowest EFE values were used to update the sampling distribution.
This procedure was repeated for 10 iterations, and the mean of the final distribution was used as the optimized action sequence.
Because the action optimization was computationally expensive and showed poor preliminary performance, which was also reflected in the quantitative results in Table~III, this baseline was limited to nine no-uncertainty cases: locations (1), (3), and (5) in Fig.~\ref{fig:env}, combined with three goal images, with one trial per case.
\endgroup

\subsection{\rev{Additional Robustness Evaluations}}
\begingroup\color{black}
The third set of experiments evaluates the proposed framework under more varied conditions: a random setting and an unseen-obstacle setting.
In the random setting, the robot was initialized at random positions and orientations, and the goal image was randomly selected from the collected observation data.
This setting was evaluated over 36 trials using the same inference configuration as the main experiment.

In the unseen-obstacle setting, additional stools were placed in the environment.
We selected two cases in which stools appeared in both the initial and goal observations and replaced the visible stools in the initial and/or goal images with stools of different colors.
This produced four conditions: unknown start only, unknown goal only, both start and goal unknown, and both start and goal known with only the unseen obstacle present.
Each of the two cases and four conditions was evaluated twice, resulting in 16 trials.

\endgroup

\section{Results}

This section presents the experimental results of the proposed framework. We begin by evaluating the policy model, examining whether it is capable of generating diverse and context-dependent action sequences. We then assess the predictive capability of the world model through imagination experiments. Finally, we demonstrate the overall navigation performance on a real mobile robot, comparing our framework with baseline methods.

\subsection{Policy Model Evaluation}

\begin{figure}[t]
    \centering
    \includegraphics[width=1.0\linewidth]{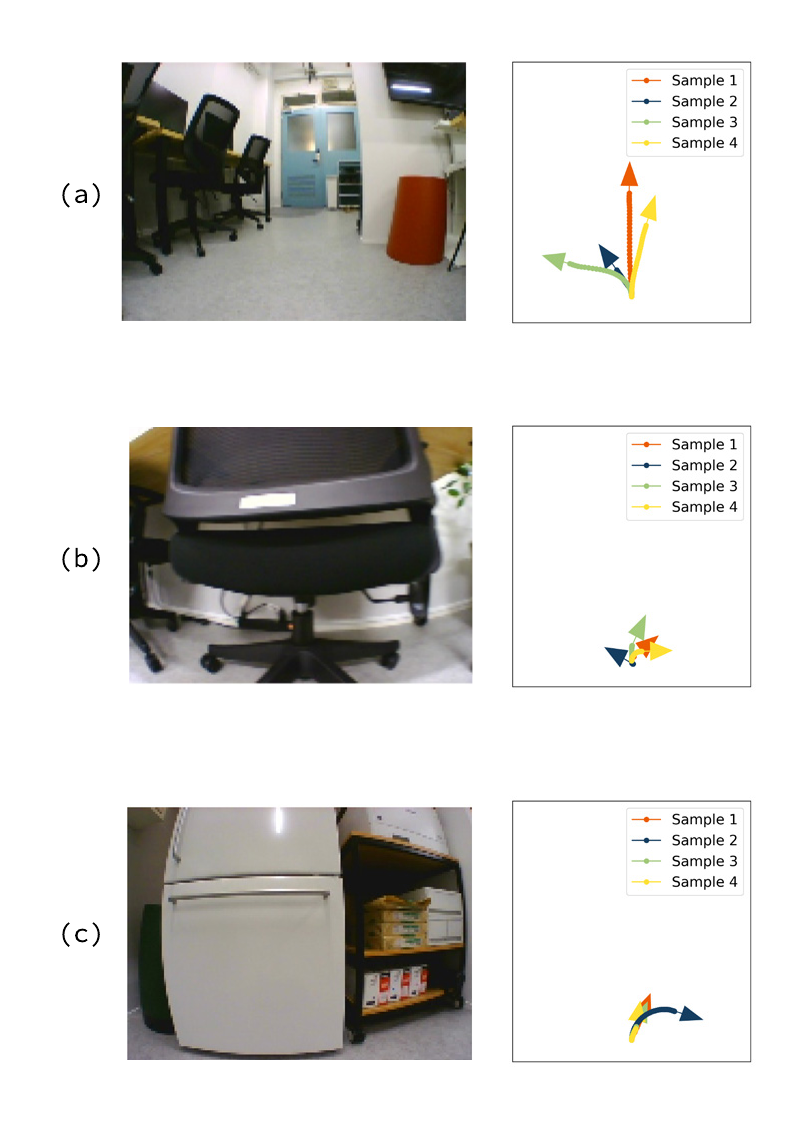}
    \caption{
    Representative action sequences generated by the diffusion policy in three scenarios: (a) clear path, (b) obstacle ahead, and (c) approaching a corner. 
    The policy adapts its action proposals to each situation and generates diverse behaviors such as forward movement and turns, illustrating its flexibility in handling different environmental contexts.
    }
    \label{fig:diffusion}
\end{figure}

We first evaluated the diffusion policy to assess its ability to generate diverse and context-dependent action sequences. Fig. \ref{fig:diffusion} shows representative cases illustrating how the policy responds under different environmental conditions.

Fig. \ref{fig:diffusion}(a) shows the case of a clear path ahead, where the corridor in front of the robot was unobstructed. The policy generated a wide variety of forward-directed actions, including driving straight ahead or turning left/right.

Fig. \ref{fig:diffusion}(b) illustrates a case in which a black desk chair was positioned directly in front of the robot.
The generated action sequences included turns to both the left and right, as well as slight forward adjustments to navigate around the obstacle.
This indicates that the policy successfully incorporated the perceived obstacle into its proposals.

Fig. \ref{fig:diffusion}(c) presents the case of approaching a room corner.
The policy produced turning actions that aligned with the room's geometry, suggesting that it can propose adaptive trajectories consistent with the layout of the environment.

Overall, these examples demonstrate that the diffusion policy is capable of flexibly generating diverse actions suited to the current situation.
While this evaluation highlights adaptability to different situations, the role of exploratory and goal-directed behaviors will be further examined in the subsequent real-world experiments.

\subsection{World Model Evaluation}
Next, we evaluated the MTRSSM, examining its predictive capability in long-horizon imagination. The model received observation inputs for the first 20 time steps and then generated predictions for 200 steps without further observations.

\begin{figure*}
    \centering
    \includegraphics[width=1.0\linewidth]{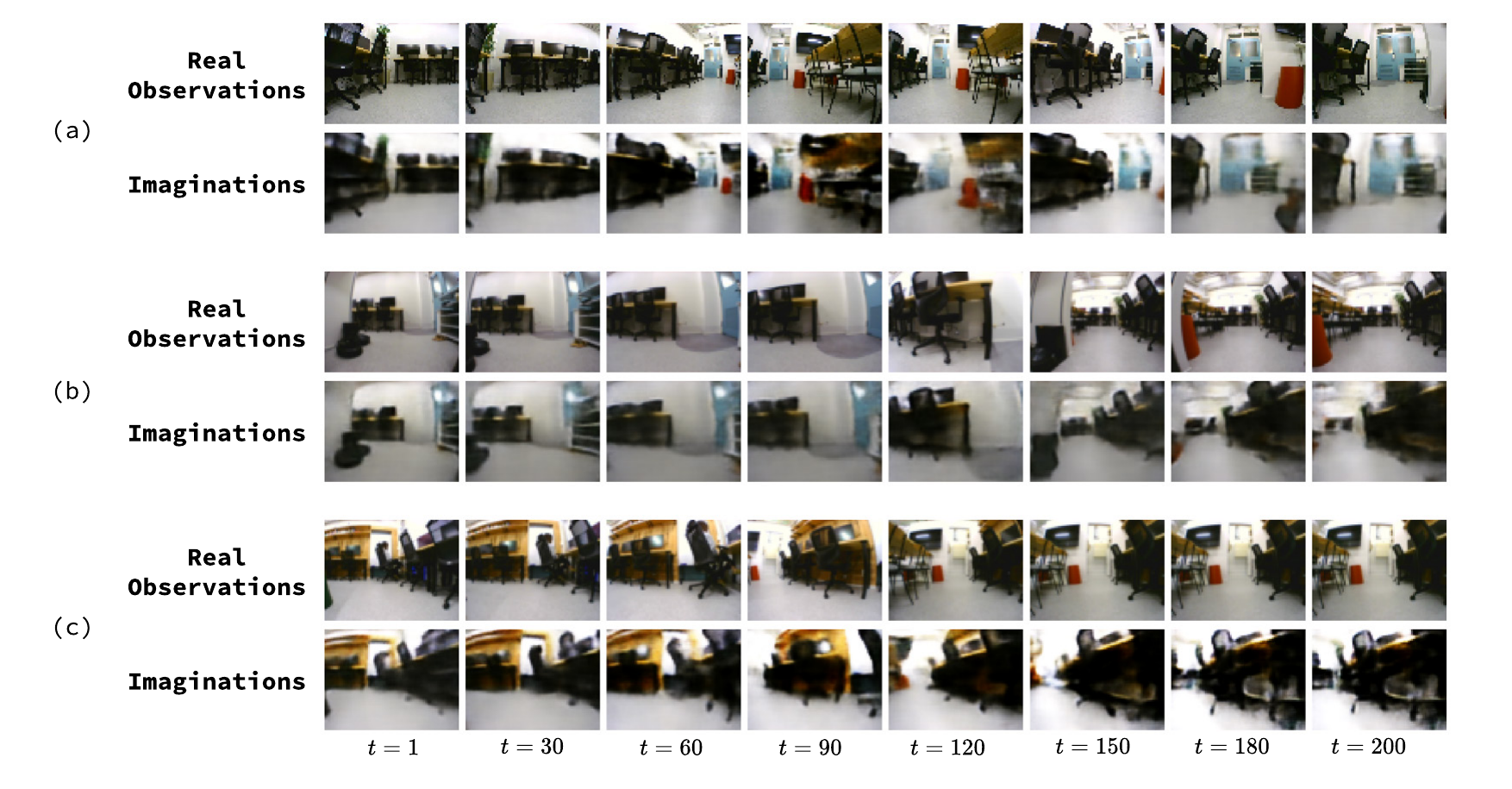}
    \caption{
    Examples of real observations and imaginations generated by the MTRSSM.
    (a, b) Successful cases. (c) Failure case.
    In the successful cases, the imaginations captured the overall scene dynamics, although subtle temporal lags appeared after time step $t = 180$.
    In the failure case, temporal lags occurred after time step $t = 120$, followed by a spatial discontinuity at time step $t = 180$.
    Despite these deviations, the imagined sequence remained coherent for more than 100 steps, demonstrating the model's long-horizon prediction capability.
    }
    \label{fig:imagine}
\end{figure*}

Figs.~\ref{fig:imagine}(a) and \ref{fig:imagine}(b) show successful \rev{cases} in which the imaginations closely matched the real observations.
Fig.~\ref{fig:imagine}(c) shows a case in which the imagination diverged from the actual observations, generating a wall-side black desk chair instead of a corridor with a red stool.
Nevertheless, the imagined sequence remained coherent for more than 100 time steps, demonstrating the model's ability to maintain long-horizon predictions.

\begin{figure*}[t]
    \centering
    \includegraphics[width=1.0\linewidth]{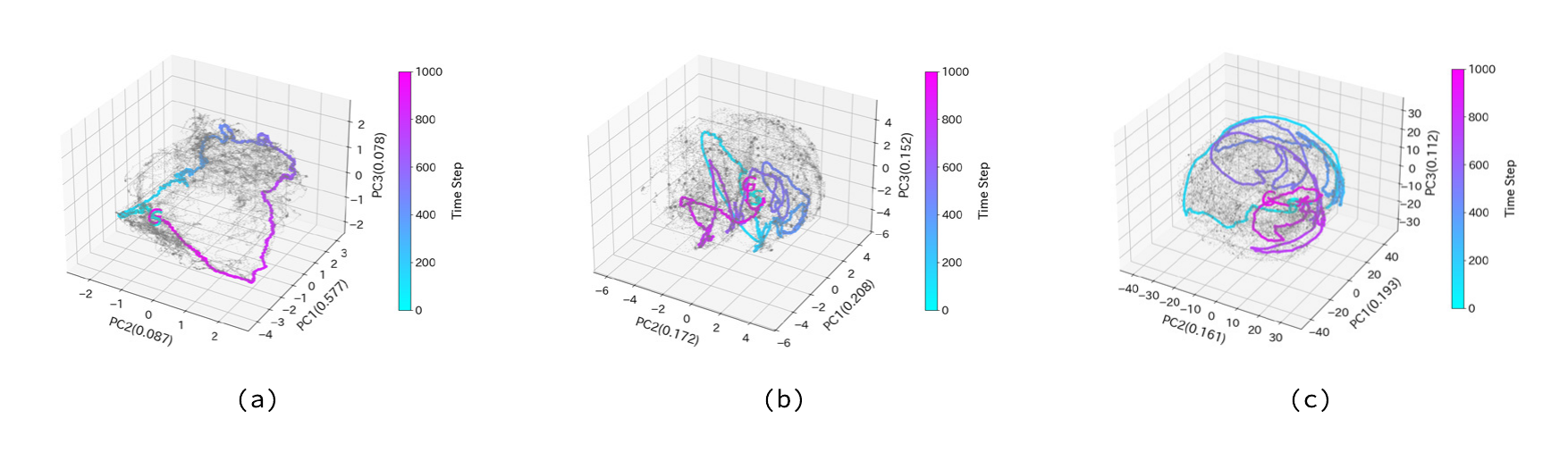}
    \caption{
    Internal state dynamics during a loop around the room.
    (a) Higher-level deterministic states $d^\mathrm{h}_t$.
    (b) Lower-level deterministic states $d^\mathrm{l}_t$.
    (c) Image features.
    For each panel, trajectories were obtained by collecting the corresponding states or features across the entire training dataset and projecting them into three dimensions using principal component analysis.
    Gray points represent all data from the training set, and the trajectory corresponding to the room loop is highlighted with a color gradient from light blue to pink, indicating temporal progression.
    Only the higher-level states $d^\mathrm{h}_t$ formed a smooth closed loop, which reflected the global room structure and demonstrated that the higher-level dynamics of the MTRSSM successfully captured the environmental geometry, enabling consistent imagination even in the absence of observation inputs.
    }
    \label{fig:state_dynamics}
\end{figure*}

Further analyzing these results, Fig.~\ref{fig:state_dynamics} illustrates the internal state dynamics during a loop around the room. Fig.~\ref{fig:state_dynamics}(a) shows the higher-level deterministic states $d_t^\mathrm{h}$, Fig.~\ref{fig:state_dynamics}(b) shows the lower-level deterministic states $d_t^\mathrm{l}$, and Fig.~\ref{fig:state_dynamics}(c) shows the image features.
For each panel, the trajectories were obtained by collecting the corresponding states or features across the entire training dataset and projecting them into three dimensions, using principal component analysis.
In the visualization, all data are shown in gray, and the trajectory corresponding to the room loop is highlighted with a time-encoded colormap.
Only the higher-level states $d_t^\mathrm{h}$ formed a smooth closed loop, which represents the global room structure and acts as an attractor in the state space.
This indicates that the higher-level dynamics of the MTRSSM successfully captured the environmental geometry, enabling consistent imagination even in the absence of observation inputs.
\rev{This connection between the latent states and the environment was not imposed by explicit position labels, but emerged through world-model training under the multiple-timescale structure.
Because long-horizon reconstruction requires both global environmental information and short-term visual details, the multiple-timescale structure of MTRSSM encourages a functional separation across hierarchy: the slowly changing higher-level state retains global environmental dynamics and self-location-related information, whereas the rapidly changing lower-level state represents short-term visual details.}

\rev{
To further examine the effect of temporal hierarchy, we compared the prediction quality of MTRSSM with RSSM variants, including multi-layer and larger RSSM baselines, as shown in Table~\ref{tab:world_model_size}.
MTRSSM achieved the best long-horizon imagination quality among these models, while the multi-layer and larger RSSM baselines did not reach the same level as MTRSSM.
These results suggest that the multiple-timescale structure, rather than model depth or capacity alone, contributes to stable long-horizon imagination for EFE computation.
}
\begin{table*}[t]
\centering
\begingroup\color{black}
\caption{
World-model size and 200-step prediction quality.
}
\label{tab:world_model_size}
\begin{tabular}{lccccc}
\toprule
\multirow{2}{*}{Model} & \multirow{2}{*}{Parameters} & \multicolumn{2}{c}{Reconstruction} & \multicolumn{2}{c}{Imagination} \\
 & & LPIPS $\downarrow$ & SSIM $\uparrow$ & LPIPS $\downarrow$ & SSIM $\uparrow$ \\
\midrule
MTRSSM & 164,608 & $\mathbf{0.074 \pm 0.004}$ & $\mathbf{0.906 \pm 0.002}$ & $\mathbf{0.204 \pm 0.033}$ & $\mathbf{0.503 \pm 0.060}$ \\
RSSM & 246,432 & $0.092 \pm 0.002$ & $0.892 \pm 0.002$ & $0.277 \pm 0.009$ & $0.392 \pm 0.041$ \\
2-layer RSSM & 229,280 & $0.086 \pm 0.003$ & $0.897 \pm 0.002$ & $0.269 \pm 0.033$ & $0.411 \pm 0.046$ \\
3-layer RSSM & 298,464 & $0.096 \pm 0.004$ & $0.890 \pm 0.005$ & $0.318 \pm 0.013$ & $0.360 \pm 0.024$ \\
RSSM (larger) & 887,104 & $0.096 \pm 0.006$ & $0.892 \pm 0.005$  & $0.235 \pm 0.010$ & $0.486 \pm 0.032$ \\
\bottomrule
\end{tabular}
\endgroup
\end{table*}

Overall, these results demonstrate that the MTRSSM can leverage its hierarchical structure to generate coherent long-horizon predictions.
This predictive capability is essential for evaluating candidate action sequences in AIF and will be combined with the diffusion policy in the subsequent real-world experiments.

\subsection{Real-World Experiments}
Finally, we evaluated the proposed deep AIF framework in real-world navigation tasks, using the setup described in Sec.~\ref{sec:navtest}.

\rev{
Table~\ref{tab:baseline_results} compares the proposed framework with the Only Extrinsic baseline and external pretrained navigation baselines.
Overall, our framework achieved a success rate of 75\%, outperforming Only Extrinsic (53\%), FlowNav (50\%), and NoMaD (8.33\%). 
We also achieved fewer collisions overall (0.806) than Only Extrinsic (1.000), FlowNav (1.306), and NoMaD (2.861).
The improvement over Only Extrinsic was especially clear in exploration-demanding scenarios (78\% vs. 28\%), indicating that epistemic value helped reduce self-localization uncertainty before goal-directed navigation.
The lower performance of FlowNav and NoMaD suggests that directly applying pretrained navigation policies to our environment was difficult.
Because these methods rely on visual localization, once the robot deviated from the topological map, localization became unreliable, making recovery difficult.
In addition, fine obstacles such as chair and desk legs may have contributed to increased collisions.
Moreover, NoMaD also showed less reliable goal-conditioned behavior in our setting, which may explain its low success rate.
In contrast, our framework updates latent states from observation histories using a world model trained in the experimental environment, enabling EFE-based recovery toward the goal.
}

\begin{table}[t]
\centering
\begingroup\color{black}
\caption{
Navigation baseline comparison results.
}
\label{tab:baseline_results}
\setlength{\tabcolsep}{3pt}
\renewcommand{\arraystretch}{1.05}
\begin{tabular}{lccc|ccc}
\toprule
\multirow{2}{*}{Method} & \multicolumn{3}{c|}{Success Rate (\%)} & \multicolumn{3}{c}{Collisions} \\
 & Overall & Exp & NoExp & Overall & Exp & NoExp \\
\midrule
Ours           & \textbf{75} & \textbf{78} & 72 & \textbf{0.806} & \textbf{0.778} & 0.833 \\
Only Ext.      & 53 & 28 & \textbf{78} & 1.000 & 1.667 & \textbf{0.333} \\
FlowNav        & 50 & 50 & 50 & 1.306 & 1.110 & 1.500 \\
NoMaD          & 8.33 & 5.55 & 11.1 & 2.861 & 2.778 & 2.944 \\
\bottomrule
\end{tabular}
\par\vspace{1mm}
\parbox{0.75\linewidth}{\footnotesize\textit{Note:} Exp and NoExp denote cases with and without exploration demand, respectively. Collisions denotes the average number of collisions per trial.}
\endgroup
\end{table}

\rev{Table~\ref{tab:ablation_results} summarizes the policy and world-model ablations.
Replacing the MTRSSM with RSSM reduced the overall success rate from 75\% to 64\%.
The contribution of MTRSSM was particularly important in exploration-demanding cases, where perceptual aliasing made self-localization difficult.
By maintaining information obtained through exploration over longer horizons, the MTRSSM supported more reliable imagination for EFE computation and helped the proposed framework maintain high success rates even in these challenging cases.
Replacing the diffusion policy with the Gaussian Sampling Policy reduced the success rate to 36.1\%, and CEM also showed lower performance with more collisions in the restricted no-exploration setting.
These results indicate that the diffusion policy served as an effective data-driven action prior, enabling more reliable candidate generation than simple Gaussian sampling or direct online optimization.
}

\begin{table}[t]
\centering
\begingroup\color{black}
\caption{
Component ablation results.
}
\label{tab:ablation_results}
\setlength{\tabcolsep}{4pt}
\begin{tabular}{lccc|ccc}
\toprule
\multirow{2}{*}{Method} & \multicolumn{3}{c|}{Success Rate (\%)} & \multicolumn{3}{c}{Collisions} \\
 & Overall & Exp & NoExp & Overall & Exp & NoExp \\
\midrule
Ours                     & \textbf{75} & \textbf{78} & \textbf{72} & \textbf{0.806} & \textbf{0.778} & 0.833 \\
RSSM                     & 64 & 61 & 67 & \textbf{0.806} & 1.056 & \textbf{0.556} \\
GSP & 36.1 & 44.4 & 27.8 & 0.944 & 0.889 & 1.000 \\
CEM (9 trials)           & 0 & -- & 0 & 3.778 & -- & 3.778 \\
\bottomrule
\end{tabular}
\par\vspace{1mm}
\parbox{0.9\linewidth}{\footnotesize\textit{Note:} Exp and NoExp denote cases with and without exploration demand, respectively. Collisions denotes the average number of collisions per trial.}
\endgroup
\end{table}

\rev{Table~\ref{tab:robustness_results} reports the additional robustness evaluations.
In the random setting, the proposed framework achieved a 69.4\% success rate, suggesting reasonable generalization to randomized initial and goal settings within the training environment, although the performance slightly decreased from the main evaluation.
Possible factors behind the reduced performance are discussed later in the failure-case analysis.
In the unseen-obstacle setting, the success rate dropped to 37.5\% and collisions increased to 1.563.
This decrease suggests that observations outside the training distribution can affect both latent-state prediction in the world model and feature-based goal evaluation, making navigation more difficult under unseen visual conditions.}

These contributions are further illustrated by the following qualitative examples, which show how epistemic and extrinsic values guide action selection in practice.

\begin{table}[t]
\centering
\begingroup\color{black}
\caption{
Robustness evaluation results.
}
\label{tab:robustness_results}
\begin{tabular}{lccc}
\toprule
Setting & Trials & Success Rate (\%) & Collisions \\
\midrule
Main setting & 36 & 75.0 & 0.806 \\
Random setting & 36 & 69.4 & 0.694 \\
Unseen-obstacle setting & 16 & 37.5 & 1.563 \\
\bottomrule
\end{tabular}
\par\vspace{1mm}
\parbox{0.9\linewidth}{\footnotesize\textit{Note:} Collisions denotes the average number of collisions per trial.}
\endgroup
\end{table}

\begin{figure*}[t]
    \centering
    \includegraphics[width=1.0\linewidth]{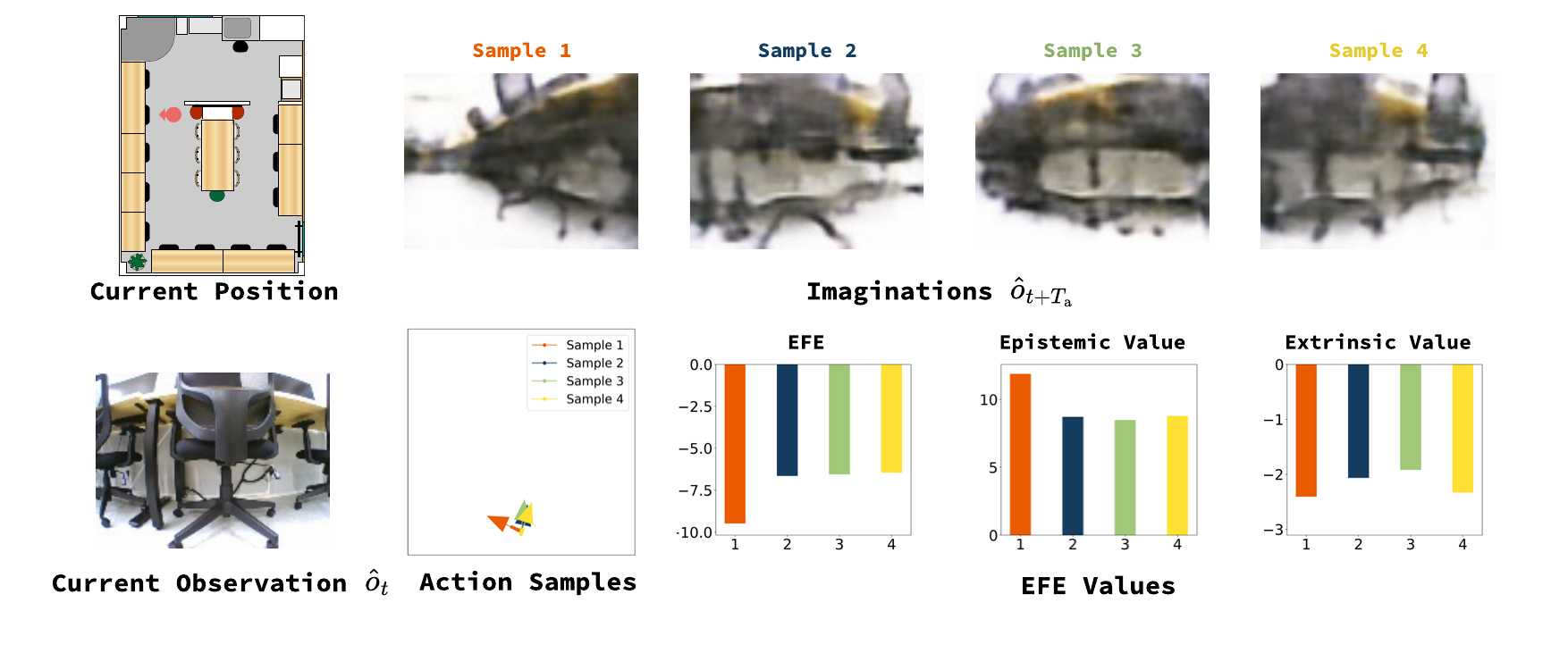}
    \caption{
    Action selection during early-stage navigation.
    The robot evaluates candidate actions, such as staying in place or turning, while the current observation---showing a black desk chair that appears at multiple locations---is insufficient for precise localization.
    Imagined observations corresponding to each action suggest that turning reveals additional environmental information.
    In this situation, the EFE is dominated by the epistemic value, leading the robot to select the turning action that reduces uncertainty and improves localization.
    }
    \label{fig:robot1}
\end{figure*}

\begin{figure*}[t]
    \centering
    \includegraphics[width=1.0\linewidth]{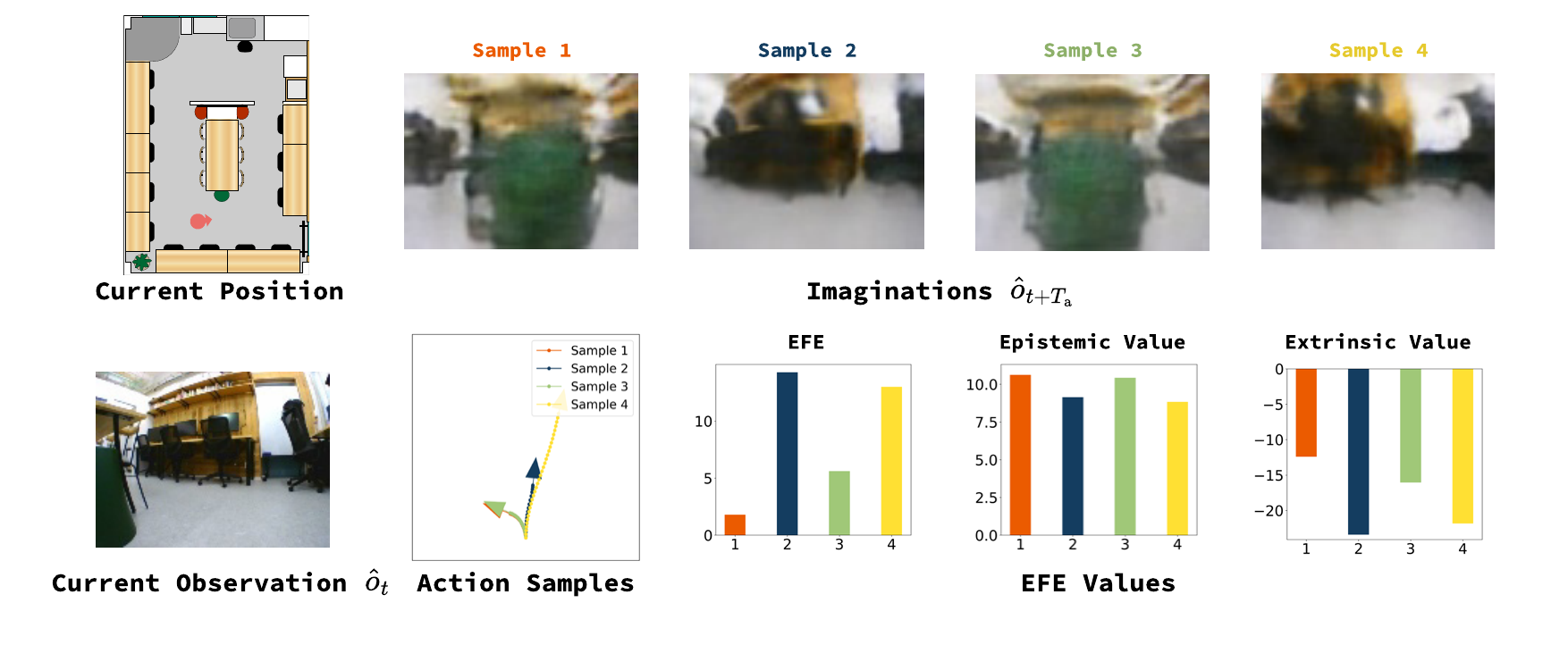}
    \caption{
    Action selection near the goal.
    The robot evaluates candidate actions, some of which would pass the goal while others approach it.
    Imagined observations corresponding to goal-approaching actions closely match the goal image.
    In this situation, the EFE is dominated by the extrinsic value, leading the robot to select actions that move it closer to the goal.
    }
    \label{fig:robot2}
\end{figure*}

\rev{Figs}.~\ref{fig:robot1} and \ref{fig:robot2} provide qualitative examples of action selection.
Here, we focus on a trial whose initial and goal observations correspond to images (4) and (5) in Fig.~\ref{fig:env}, respectively.
Fig.~\ref{fig:robot1} shows the early stage of the trial, whereas Fig.~\ref{fig:robot2} shows the later stage near the goal.
These examples illustrate how the robot selected actions at different phases of navigation in the trial.

In Fig.~\ref{fig:robot1}, the robot initially faced a black desk chair, and because similar chairs were placed throughout the environment, the current observation alone was insufficient for reliable self-localization.
Candidate actions included staying in place as well as turning in place.
The imagined observations at the last time step revealed that the turning action (Sample 1 in the figure) would expose additional information beyond the line of chairs along the wall.
Our framework selected this turning action because the EFE assigned it a high epistemic value, reflecting the potential information gain.
This illustrates how epistemic considerations guide the robot toward exploratory behavior, enabling it to reduce uncertainty and improve subsequent localization.

In contrast to the early stage shown in Fig.~\ref{fig:robot1}, Fig.~\ref{fig:robot2} illustrates the later stage of the same trial, when the robot was already near the goal.
In the current observation, a green stool---also present in the goal observation---appears on the left side, providing a clear visual cue.
At this point, some candidate actions would continue past the goal, whereas others would turn toward and approach it.
The imagined observations for the goal-approaching actions (Samples 1 and 3) closely matched the goal observation.
In this case, the EFE was dominated by the extrinsic value, favoring actions that brought the robot closer to the goal.

\begin{figure*}[t]
    \centering
    \includegraphics[width=1.0\linewidth]{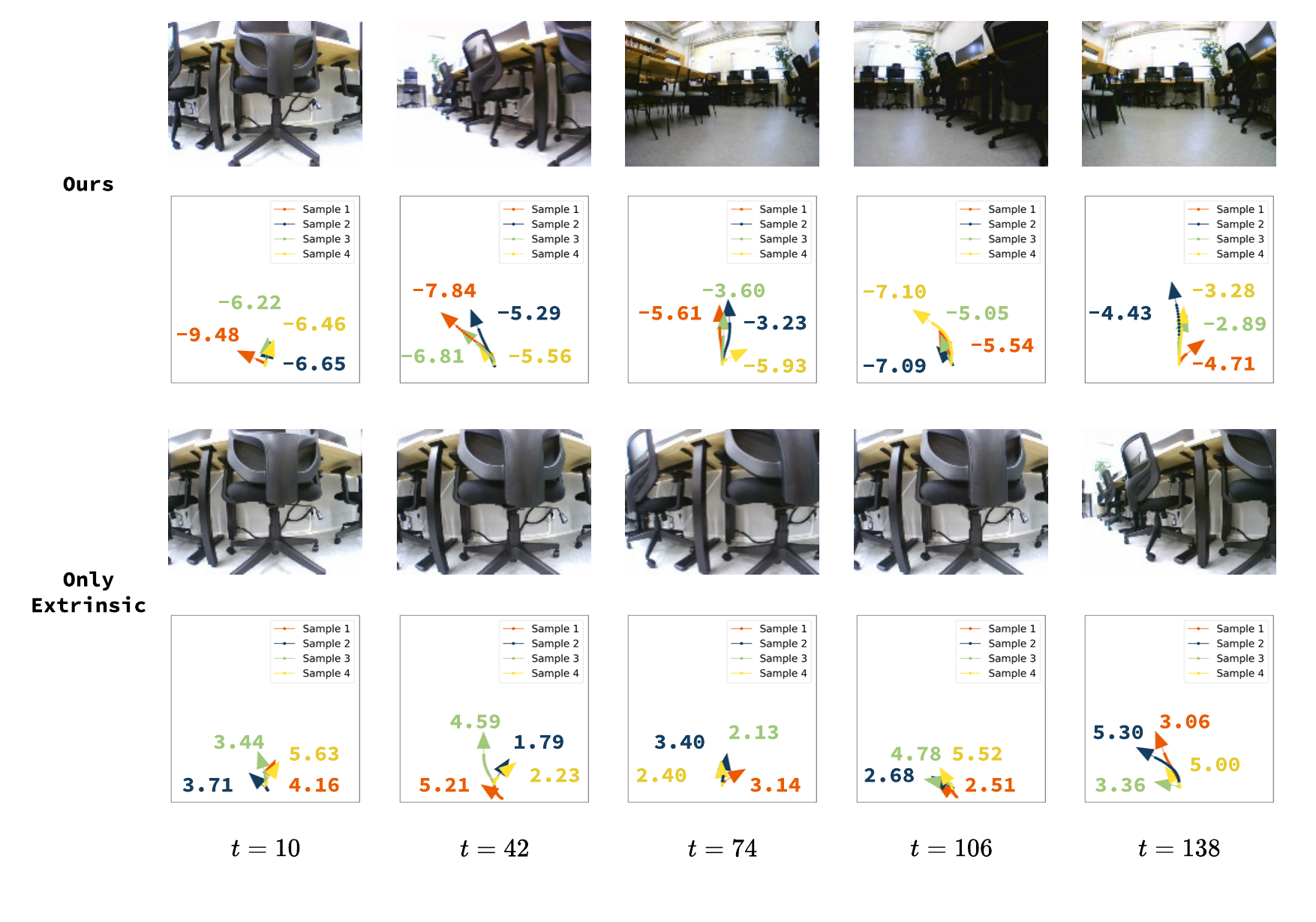}
    \caption{
    Comparison of action selection between our deep AIF framework (upper row) and an Only Extrinsic baseline (lower row) in an exploration scenario.
    Our framework demonstrates active exploration, acquiring diverse observations over time, whereas the baseline largely remains stationary.
    The first action in our framework corresponds to the example in Fig.~\ref{fig:robot1}, highlighting the role of the epistemic value in driving exploratory behavior.
    }
    \label{fig:robot3}
\end{figure*}

Fig.~\ref{fig:robot3} compares our framework with the Only Extrinsic baseline in the same scenario that required exploration.
Note that the first situation ($t = 10$) in our framework corresponds to the example shown in Fig.~\ref{fig:robot1}.
The figure presents the sampled candidate actions together with their EFE values.
The action with the lowest EFE was selected and executed, leading to a change in the robot's observations.
The upper row (our framework) demonstrates active exploration, where the robot obtained increasingly diverse observations over time, whereas the lower row (Only Extrinsic) shows the robot largely remaining in place without gaining new information.
These results indicate that incorporating epistemic value into the EFE formulation encouraged exploratory actions that produced new observations, thereby enabling the robot to resolve uncertainty and achieve self-localization, whereas the baseline relying solely on extrinsic value failed to do so.

\rev{We further examined whether the epistemic value reflected uncertainty rather than visual novelty alone.
We compared non-uncertain observations, corresponding to patterns (1), (3), and (5) in Fig.~\ref{fig:env}(c), with perceptually aliased uncertain observations, corresponding to patterns (4), (6), and (8) in Fig.~\ref{fig:env}(c).
After initializing the world-model state with each image, we applied the same 180-degree in-place rotation action.
The uncertain cases showed a higher mean epistemic value (13.291 vs. 12.498) and a larger absolute mean change (0.212 vs. 0.094), suggesting that the epistemic value reflects observation uncertainty and its action-dependent change.}

\rev{Finally, we analyzed the main factors underlying the failure cases.
Most failures were associated with world-model prediction errors near visually similar or close-up objects, such as multiple stools or black desk chairs.
In these situations, small errors in imagined observations could affect the extrinsic-value evaluation, especially when the robot turned close to objects along the wall side.
Additional factors were also observed in some cases.
The feature encoder $f(\cdot)$ did not always provide a reliable goal-proximity estimate under perceptual aliasing or difficult goal images, and the non-goal-conditioned policy sometimes generated actions consistent with the training-data statistics rather than recovery actions after unfavorable movements.}

\section{Discussion}
The central finding of this study is that AIF can effectively unify exploration and goal-directed navigation in a real-world setting by minimizing EFE.
Although previous research has tended to emphasize either the epistemic value to drive exploration \cite{detinguy2024, RobotnavasAIF} or the extrinsic value to pursue navigation objectives \cite{catal2020, simAIF}, few studies have simultaneously integrated both aspects in physical robots\cite{Bio}.
As shown in Table~\ref{tab:baseline_results}, our framework achieved higher success rates, particularly in exploration-demanding tasks, clearly indicating the benefit of incorporating the epistemic value.
This finding highlights the potential of AIF as a principled framework for navigation under partial observability and perceptual aliasing.

A key contribution of this work lies in its integration of a diffusion policy into the AIF framework. Prior studies have shown that diffusion policies are capable of generating diverse and adaptive action sequences in navigation tasks \cite{NoMaD, LDP, DD, NavDP}.
However, they often rely on additional modules such as high-level planners or manual switching mechanisms to balance exploration and navigation.
For example, NoMaD~\cite{NoMaD} demonstrated the flexibility of diffusion-based action generation but still required external guidance to select between exploratory and goal-directed behaviors.
In contrast, our formulation embeds the diffusion policy within the AIF framework, enabling exploratory trajectories to be considered naturally alongside goal-directed ones.
This effect is visible in Fig.~\ref{fig:robot1}, where the epistemic value drove the robot to turn and expose new observations, thereby reducing localization uncertainty.
\rev{Another practical advantage is that EFE is used as a planning-time action-selection criterion rather than as a training reward for a fixed task.
This differs from curiosity-driven reinforcement learning, where the policy is trained for a task-specific reward objective that includes intrinsic motivation.
As a result, different goal images can be specified after training while reusing the same learned world model and policy prior.
}

Equally important is the role of the MTRSSM.
Recurrent world models such as RSSM~\cite{masteringatari} have been widely adopted for imagination-based planning, yet they often suffer from error accumulation in long-horizon predictions.
As shown in Fig.~\ref{fig:state_dynamics}, only the higher-level deterministic states of the MTRSSM formed smooth closed loops, reflecting the global room structure.
This indicates that MTRSSM successfully captured slow-varying dynamics, enabling robust long-horizon imagination.
\rev{This interpretation is further supported by the prediction-error analysis in Table~\ref{tab:world_model_size}, where MTRSSM achieved better long-horizon imagination quality than RSSM.
}
Such stability is essential for reliable estimation of epistemic and extrinsic values.

The integration of diffusion policy and MTRSSM within AIF reveals a synergistic effect.
Although diffusion policy provides a broad set of candidate actions, MTRSSM supplies reliable long-horizon predictions that enable these candidates to be meaningfully evaluated. The EFE formulation then acts as a unifying criterion selecting actions that balance epistemic and extrinsic considerations. 
\rev{To further examine this advantage, we compared our diffusion-policy-based action generation with CEM, a commonly used optimization method in traditional world-model planning frameworks \cite{sekar2020planning}.
Because CEM optimizes actions from scratch through repeated EFE evaluations, it showed lower success and substantially higher inference cost than the proposed framework.
This result highlights the advantage of combining a learned action prior with EFE-based action selection.}

From a broader perspective, the proposed framework contributes to extending the scalability of AIF in real-world robotics.
Previous applications of AIF in navigation have largely been confined to simulation environments~\cite{detinguy2024, simAIF}.
Even when deployed on real robots, these studies often addressed simplified tasks or considered exploration and navigation in isolation~\cite{RobotnavasAIF, Bio}.
By leveraging advances in generative modeling, the present work demonstrates that AIF can be scaled to more complex and uncertain real-world scenarios.
With its ability to flexibly generate actions, perform stable long-horizon imagination, and balance epistemic and extrinsic values, AIF represents a competitive alternative to traditional methods. 
In particular, our framework provides a unified mechanism that does not require explicit mapping as in SLAM-based navigation\cite{SLAM, Kadian2020, Gervet2023} or handcrafted planners\cite{NoMaD, ViNT}.
\rev{SLAM mainly addresses mapping and localization, whereas action selection typically requires additional planning modules.
In our framework, the world model plays a localization-related role through observation prediction, while action selection is formulated directly by EFE minimization.}

Nevertheless, several limitations must be acknowledged.
First, the experimental environment was limited to a single indoor room, which limited the diversity of the conditions tested.
\rev{Although the random and unseen-obstacle evaluations in Table~\ref{tab:robustness_results} partially address this issue, the unseen-obstacle result shows that adaptation to substantially changed environments remains limited.}
Second, although MTRSSM reduced long-horizon prediction errors compared with RSSM, deviations such as spatial discontinuity still occurred (Fig.~\ref{fig:imagine}(c)).
\rev{Third, real-time feasibility remains a practical limitation because EFE computation over candidate action sequences introduces non-negligible inference latency.
Improving the efficiency of EFE evaluation is therefore an important direction for future work.}
\rev{Fourth, the precision for the extrinsic term was heuristically scheduled as in (\ref{eq:precision}) for numerical stability, introducing a heuristic element into the active inference formulation.
This schedule affects navigation performance because the relative scale of epistemic and extrinsic values varies across task conditions.
Automatically adapting this balance remains an open issue in AIF.
A promising direction is to improve the world model so that epistemic value more accurately reflects task-relevant uncertainty, reducing the need for manually scheduled weighting.
}


\section{Conclusion}
In this paper, we proposed a deep AIF framework for real-world navigation that incorporates a diffusion policy as the policy model and an MTRSSM as the world model.
The diffusion policy enabled the generation of diverse and context-dependent action candidates, while the MTRSSM provided stable long-horizon predictions that preserved the environmental structure.
The integration of the policy and world models within the EFE formulation enabled the epistemic and extrinsic values to be exploited effectively, resulting in improved navigation performance compared with baseline methods.
\rev{Additional baseline comparisons, ablations, and robustness evaluations further validated the proposed design while revealing the remaining challenge of adapting to unseen visual changes.
}
These findings demonstrate that recent advances in deep generative modeling can substantially enhance the scalability of AIF in real-world robotic systems.

Looking ahead, we identify three promising directions for future research.
First, incorporating natural language instructions as goal specifications might enable more flexible and intuitive navigation.
Second, leveraging pretrained foundation models might facilitate adaptation to entirely unseen environments.
Finally, extending the framework to manipulation and integrating it with navigation might open the way for deep AIF-based mobile manipulation in real-world robotics.

\bibliographystyle{IEEEtran}
\bibliography{reference}
\newpage

\end{document}